\documentclass[letterpaper, 10 pt, journal, twoside]{IEEEtran}

\usepackage{graphics} 
\usepackage{epsfig} 
\usepackage{amsmath} 
\usepackage{amssymb}  
\usepackage{siunitx}
\usepackage{soul, color} 
\usepackage[percent]{overpic}
\makeatletter\newcommand{\manuallabel}[2]{\def\@currentlabel{#2}\label{#1}}\makeatother
\usepackage{hyperref}

\usepackage{fancyhdr}
\begin{document}
\title{
Robust Footstep Planning and LQR Control for Dynamic Quadrupedal Locomotion
}

\author{Guiyang Xin$^{1}$, Songyan Xin$^{1}$, Oguzhan Cebe$^{1}$, Mathew Jose Pollayil$^{2}$, Franco Angelini$^{2}$, \\Manolo Garabini$^{2}$, Sethu Vijayakumar$^{{1}, {3}}$, Michael Mistry$^{1}$
\thanks{Manuscript received: October, 15, 2020; Revised January, 16, 2021; Accepted February, 25, 2021.}
\thanks{This paper was recommended for publication by Editor Abderrahmane Kheddar upon evaluation of the Associate Editor and Reviewers' comments.
This work was supported by the following grants: EPSRC UK
RAI Hubs NCNR (EP/R02572X/1), ORCA (EP/R026173/1), EU Horizon 2020 THING (ICT-27-2017 780883) and NI (ICT-47-2020 101016970).} 
\thanks{$^{1}$Guiyang Xin, Songyan Xin, Oguzhan Cebe, Sethu Vijayakumar and Michael Mistry
are with the School of Informatics, Institute of Perception, Action and Behaviour, University of Edinburgh, EH8 9AB, 10 Crichton Street, Edinburgh, United Kingdom
        {\tt\small guiyang.xin@ed.ac.uk}}%
\thanks{$^{2}$Mathew Jose Pollayil, Franco Angelini and Manolo Garabini are with the Department of Information Engineering of the University of Pisa and with the Research Center “E. Piaggio” of the University of Pisa, Italy
        {\tt\small mathewjose.pollayil@phd.unipi.it}}%
\thanks{$^{3}$The author is a visiting researcher at the Shenzhen Institute for ArtificialIntelligence and Robotics for Society (AIRS).}%
\thanks{Digital Object Identifier (DOI): see top of this page.}
}

\markboth{IEEE Robotics and Automation Letters. Preprint Version. Accepted February, 2021}
{Xin \MakeLowercase{\textit{et al.}}: Robust Footstep Planning and LQR Control for Dynamic Quadrupedal Locomotion} 


\maketitle
\thispagestyle{fancy}

\begin{abstract}

In this paper, we aim to improve the robustness of dynamic quadrupedal locomotion through two aspects: 1) fast model predictive foothold planning, and 2) applying LQR to projected inverse dynamic control for robust motion tracking. In our proposed planning and control framework, foothold plans are updated at 400 Hz considering the current robot state and an LQR controller generates optimal feedback gains for motion tracking. The LQR optimal gain matrix with non-zero off-diagonal elements leverages the coupling of dynamics to compensate for system underactuation. Meanwhile, the projected inverse dynamic control complements the LQR to satisfy inequality constraints. In addition to these contributions, we show robustness of our control framework to unmodeled adaptive feet. Experiments on the quadruped  ANYmal demonstrate the effectiveness of the proposed method for robust dynamic locomotion given external disturbances and environmental uncertainties.
\end{abstract}

\begin{IEEEkeywords}
Legged Robots, Whole-Body Motion Planning and Control, Motion Control
\end{IEEEkeywords}
\section{INTRODUCTION}
\label{section:introduction}

\IEEEPARstart{L}{egged} robots have evolved quickly in recent years. Although there are robots, such as Spot from Boston Dynamics, which have been deployed in real industrial scenarios, researchers continue to explore novel techniques to improve locomotion performance. A popular technique is the staged approach which divides the larger problem into sub-problems and chains them together. Typically the pipeline is composed of state estimation, planning and control, which may be running at different frequencies. The motion planner typically runs at a slower frequency comparing to controller due to model nonlinearities and long planning horizons. The lower-level feedback controller runs at a higher frequency to resist model discrepancies and external disturbances.  After years of evolution, optimization becomes the core approach for motion planning and control of legged robots. 

\begin{figure}[t!]
\centering
 \includegraphics[width=0.8\linewidth]{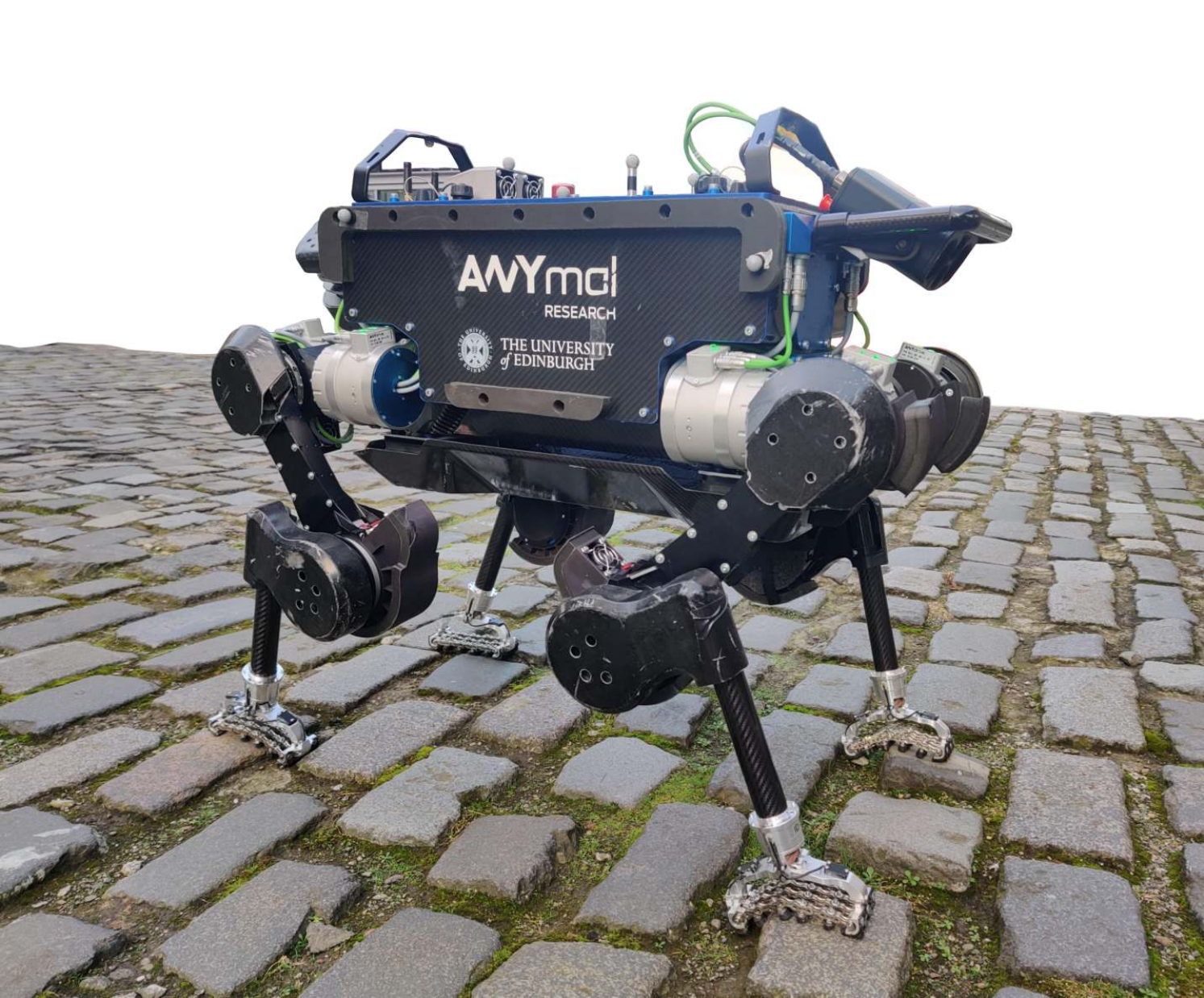}
 \caption{ANYmal with adaptive feet stepping on rough terrain.}
 \label{f:fig1}
\end{figure}

\subsection{Related planning methods}
\label{subsection:related_planning_methods}
Legged robot motion planning is a trade off between several criteria: formulation generality, model complexity, the planning horizon and computational efficiency. 
While the goal is to maximize all at once, this is not realistic given current available computational resources. As a result, different design choices lead to different formulations.

To generate motions in a more general and automated fashion, trajectory optimization (TO) has been used. 
In \cite{winkler2017fast}, a Zero Moment Point (ZMP)-based TO formulation is presented to optimize body motion, footholds and center of pressure simultaneously. It can generate different motion plans with multiple steps in less than a second.
In a later work \cite{winkler2018gait}, a phase-based TO formulation is proposed to automatically determine the gait-sequence, step timings, footholds, body motion and contact forces. Motion for multiple steps can be still generated in few seconds. 
In these two works, the TO formulations are both extremely versatile in terms of motion types that can be generated, however, online Model Predictive Control (MPC) has not been demonstrated yet. 

A linearized, single rigid-body model has been proposed \cite{di2018dynamic}\cite{kim2019highly} to formulate the ground reaction force as a QP optimization problem and which can be solved in an MPC fashion. 
In both works, the footstep locations are provided by simple heuristics. 
Online TO based on a nonlinear single rigid-body model has been given in \cite{cebe2020online}, and can generate stable dynamic motion for quadruped robots based on a given contact sequence.
A whole-body dynamic model has been considered in 
\cite{koenemann2015whole}\cite{bjelonic2020whole} to generate robot motion in a MPC fashion. Crocoddyl \cite{mastalli2020crocoddyl} improves the computation speed further more. A frequency-aware MPC is proposed in \cite{8968251} to deal with the bandwidth limitation problem for real hardware. In those four works, the contact planning problem has been decoupled from the whole-body motion planning problem.

\begin{figure*}[hbt!]
\centering
 \includegraphics[width=1\linewidth]{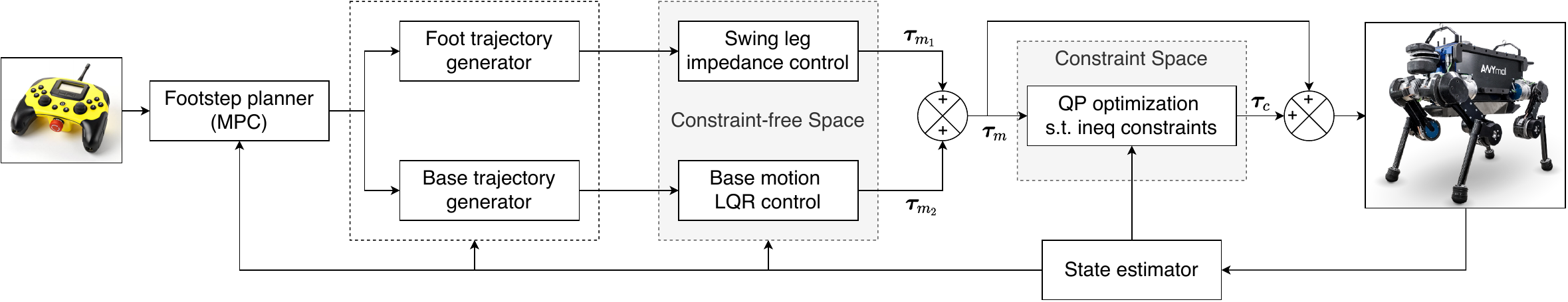}
 \caption{Control framework block diagram. All the modules are running at 400 Hz. The joystick sends desired walking velocities. The MPC generates desired ZMPs. The ZMPs are mapped to foot placements which generate swing foot trajectories by interpolation. The desired base trajectory is generated based on desired ZMPs. An LQR and two impedance controllers are employed to track desired base trajectory and swing foot trajectories in the constraint-free space. Constraints, such as torque limits and friction cone, are satisfied in the constraint space.}
 \label{f:framework}
\end{figure*}

Footstep optimization on biped robots has been proposed in \cite{faraji2014versatile}
\cite{feng2016robust}.
An underactuated linear inverted pendulum model (LIPM) has been used to formulate the footstep optimization problem.
The idea has been extended and generalized to both biped and quadruped robots in our previous work \cite{xin2019online}.
In this work, we realize real-time footstep optimization that can be executed in a MPC fashion and test it on the hardware ANYmal.

\subsection{Related control methods}
\label{subsection:related_control_methods}

In recent years, there has been a convergence among legged robot researchers to formulate the control problem as a Quadratic Program (QP) with constraints. The problem can be further decomposed into hierarchies to coordinate multiple tasks within whole-body control \cite{Saab2013}\cite{del2015prioritized}\cite{herzog2016}. These optimization-based controllers usually rely on manually tuned diagonal feedback gain matrices. Also, these controllers only compute the best commands for the next control cycle, and therefore are not suitable for dynamic gaits with underactuation. Classical optimal control theory, such as LQR, can consider long or even infinite time horizons and generate optimal non-diagonial gain matrices exploiting dynamic coupling effects which benefit underactuated systems such as a cart-pole \cite{reist2010simulation}. 

Classical LQR does not consider any constraints except system dynamics. However, for legged robots, we have to satisfy inequality constraints such as torque limits and friction cones on contact feet. The works \cite{mason2014full}\cite{mason2016balancing} proposed to use the classical LQR controller for bipedal walking, but inequality constraints were not enforced. In this paper, we propose an LQR controller for dynamic gait control under the framework of projected inverse dynamics \cite{xin2018}\cite{xin2020optimization}. Projected inverse dynamic control enables us to control motion in a constraint-free subspace while satisfying inequality constraints in an orthogonal subspace. In our previous work, we used Cartesian impedance controllers within the constraint-free subspace to control both the base and swing legs during static walking of a quadruped robot. Here, we use LQR in the constraint-free subspace to replace the Cartesian impedance controller for base motion control and to handle the underactuation in the trotting gait.

\subsection{Contributions}
\label{subsection:contributions}

This paper focuses on improving the robustness of dynamic quadrupedal gaits. The trotting and pacing gaits of a quadruped robot will be studied and demonstrated in simulations and real experiments (see Fig. \ref{f:fig1}). The main contributions lie in the computation speed of the MPC and the optimal feedback control. As an additional contribution, our approach is shown to be valid both with the default spherical feet and the adaptive feet \cite{adaptivefeet2021} with flexible soles. The main contributions are listed as follows:

\begin{enumerate}
    \item We propose to formulate foothold planning as a QP problem subject to LIPM dynamics, which can be solved within the control cycle of $\SI{2.5}{\milli\second}$. Running re-planning at high frequency allows the robot to be responsive to disturbances and control commands. The higher the updating frequency of the MPC, the better the reactivity achieved by the robot.
    \item We use unconstrained infinite-horizon LQR to generate optimal gains for base control in order to improve the robustness of the controller and cope with underactuation. Meanwhile, we inherit the advantage of our previous projected inverse dynamic framework to satisfy the inequality constraints in an orthogonal subspace, which is different to the purely QP-based controllers \cite{Saab2013}\cite{del2015prioritized}\cite{herzog2016}.
\end{enumerate}

\subsection{Paper organization}
\label{subsection:paper_organization}

The paper is organized in accordance with the hierarchical structure of the whole system, which is shown in Fig. \ref{f:framework}. Given the desired velocity, the foothold planner plans future footsteps based on the current robot state which is explained in Section \ref{section:motion_generation}. 
Section \ref{section:lqr_for_base_control} describes the derivation of the LQR for base control. 
Simulations, experiments and discussions are given in Section \ref{section:experiments}. Finally, Section \ref{section:conclusions} draws the relevant conclusions.

\section{Motion generation}
\label{section:motion_generation}

When considering dynamic gaits such as trotting, two contact points cannot constrain all six degrees of freedom (DOF) of the floating base. The system becomes underactuated as one DOF around the support line is not directly controlled. Researchers have been using the LIPM as an abstract model for balance control in this situation. The Centre of Mass (CoM) position and velocity can be predicted by solving the forward dynamics of the passive inverted pendulum. In order to keep long term balance, the next ZMP point has to be carefully selected to capture the falling CoM. For trotting, the ZMP point always lies on the support line formed by the supporting leg pair. 
As a result, the footholds optimization problem can be transformed to a ZMP optimization problem.     

\subsection{MPC formulation}
\label{subsection:mpc_formulation}

The dynamics of the linear inverted pendulum is as follows:
\begin{equation}
\begin{split}
    &\ddot{x}_{CoM}=\frac{g}{z_{CoM}}(x_{CoM}-p_x)\\
    &\ddot{y}_{CoM}=\frac{g}{z_{CoM}}(y_{CoM}-p_y)
\end{split}
\end{equation}
where $x_{CoM}$, $y_{CoM}$ and $z_{CoM}$ are the CoM position coordinates, $p_x$ and $p_y$ are the coordinates of ZMP, $g$ represents the acceleration of gravity. Considering $z_{CoM}$ as constant, the dynamics become linear and result in the following solution:
\begin{equation}\label{e:LIPM}
\begin{split}
    &\mathbf{x}_{CoM}(t)=\mathbf{A}(t)\mathbf{x}_{CoM}^0+\mathbf{B}(t)p_x\\
    &\mathbf{y}_{CoM}(t)=\mathbf{A}(t)\mathbf{y}_{CoM}^0+\mathbf{B}(t)p_y
\end{split}
\end{equation}
where $\mathbf{x}_{CoM}=\begin{matrix}[x_{CoM} & \dot{x}_{CoM}]\end{matrix}^\top$, $\mathbf{y}_{CoM}=\begin{matrix}[y_{CoM} & \dot{y}_{CoM}]\end{matrix}^\top$, are the state vectors, and $\mathbf{x}_{CoM}^0$ and $\mathbf{y}_{CoM}^0$ are the initial state vectors. $\mathbf{A}(t)$ and $\mathbf{B}(t)$ are defined as
\begin{equation}
    \mathbf{A}(t)=\begin{bmatrix}
    \cosh(\omega t) & \omega^{-1}\sinh(\omega t)\\
    \omega \sinh(\omega t) & \cosh(\omega t)
    \end{bmatrix}
\end{equation}

\begin{equation}
    \mathbf{B}(t)=\begin{bmatrix}
    1-\cosh(\omega t)\\
    -\omega \sinh(\omega t)
    \end{bmatrix}
\end{equation}
while $\omega=\sqrt{g/z_{CoM}}$.

For a periodic trotting gait with fixed swing duration $T_s$, assuming instant switching between single support phases, the states of $N$ future steps along $x$ direction can be predicted given step duration $T_{s_i}$
\begin{equation}\label{e:evolution}
\begin{split}
    \mathbf{x}_{CoM_1} =\mathbf{A}(T_{s_1})&\mathbf{x}_{CoM_0}+\mathbf{B}(T_{s_1})p_{x_1} \\
    \mathbf{x}_{CoM_2} =\mathbf{A}(T_{s_2})&\mathbf{x}_{CoM_1}+\mathbf{B}(T_{s_2})p_{x_2} \\
    & \vdots \\
    \mathbf{x}_{CoM_N} =\mathbf{A}(T_{s_N})&\mathbf{x}_{CoM_{N-1}}+\mathbf{B}(T_{s_N})p_{x_N}
\end{split}
\end{equation}
where $\mathbf{x}_{CoM_0}$ is the state at the moment of first touchdown, which can be computed from
\begin{equation}
    \mathbf{x}_{CoM_0} = \mathbf{A}(t_0) \mathbf{x}_{CoM}^{0}+\mathbf{B}(t_0)p_{x_0}
\end{equation}
where $t_0$ is the remaining period of the current swing phase. $\mathbf{x}_{CoM}^{0}$ and $p_{x_0}$ are the current CoM state and ZMP location given by the state estimator which also runs at 400 Hz. 

Also, considering the kinematic limits of the swing feet, the following inequality constraints are enforced:

\begin{equation}\label{e:kinematicLimits}
    \begin{bmatrix}p_{x_0}-d\\-d \\ \vdots \\ -d \\ -d \end{bmatrix} \leq \begin{bmatrix}1 & 0 & \cdots & 0 & 0\\ -1 & 1 & \cdots & 0 & 0 \\ \vdots & \vdots & \vdots & \vdots & \vdots \\ 0 & 0 & \cdots & 1 & 0 \\ 0 & 0 & \cdots & -1 & 1\end{bmatrix} \begin{bmatrix}p_{x_1}\\p_{x_2} \\ \vdots \\p_{x_{N-1}} \\ p_{x_N}\end{bmatrix} \leq \begin{bmatrix}p_{x_0}+d\\ d \\ \vdots \\ d \\ d \end{bmatrix} 
\end{equation}

where $d$ is a constant value derived from kinematic reachability relative to the stance feet. Additionally, Eq. (\ref{e:kinematicLimits}) can also be used to avoid stepping into unfeasible pitches on the ground by redefining $d$.

The state along $y$ direction has the same evolution as shown in Eq. (\ref{e:evolution}). Regarding ZMPs as the system inputs, we define the cost function of the MPC as follows
\begin{equation}\label{e:MPCCost}
    \displaystyle\sum_{i=1}^{N} \frac{1}{2}[Q_i(\dot{x}_{CoM_i}-\dot{x}_{CoM_d})^2+R_i(p_{x_i}-p_{x_{i-1}})^2]
\end{equation}
where $\dot{x}_{CoM_d}$ is the desired CoM velocity in the $x$ direction, $Q_i$ and $R_i$ are weight factors. The cost function for the $y$ direction has the same form as Eq. (\ref{e:MPCCost}). The MPC is formulated as a QP minimizing Eq. (\ref{e:MPCCost}) subject to Eq. (\ref{e:evolution}) and Eq. (\ref{e:kinematicLimits}). Solving the QP results in the optimal ZMPs for the future $N$ steps $\mathbf{p}_x^*=\begin{matrix}[p_{x_1}^* & p_{x_2}^* &\dots & p_{x_N}^*]\end{matrix}^\top$. 

Similarly, solving another QP for $y$ direction yields the coordinate ${\mathbf{p}_y^*=\begin{matrix}[p_{y_1}^* & p_{y_2}^* &\dots & p_{y_N}^*]\end{matrix}^\top}$ for the optimal ZMPs in this direction. It should be noted that the cost function for $y$ direction is slightly different to Eq. (\ref{e:MPCCost}), which is as follows
\begin{equation}\label{e:MPCCostY}
    \displaystyle\sum_{i=1}^{N} \frac{1}{2}[Q_i(\dot{y}_{CoM_i}-\dot{y}_{CoM_d})^2+R_i(p_{y_i}-p_{y_{i-1}}-s(-1)^ir_y)^2]
\end{equation}
where $r_y$ is a constant distance between right and left ZMPs. $r_y \neq 0$ for pacing gait to avoid self-collision while $r_y = 0$ for trotting gait. $s$ indicates the support phase the robot is in, $s = 1$ for left support and $s= -1$ for right support.

We only use the first pair $\mathbf{p}_1^{*}=(p_{x_1}^* \quad p_{y_1}^*)$ to generate the swing trajectory. Since the MPC is running in the same loop of controller, the position $\mathbf{p}_1^{*}=(p_{x_1}^* \quad p_{y_1}^*)$ keeps updating during a swing phase given the updated CoM state $(\mathbf{x}_{CoM}^0$ \enskip $\mathbf{y}_{CoM}^0)$ and desired CoM velocity $(\dot{x}_{CoM_d}$ \enskip $\dot{y}_{CoM_d})$.

\subsection{Reference trajectories of trotting gait}
\label{subsection:reference_trajectories}

\begin{figure}[t!]
\centering
 \includegraphics[width=1\linewidth]{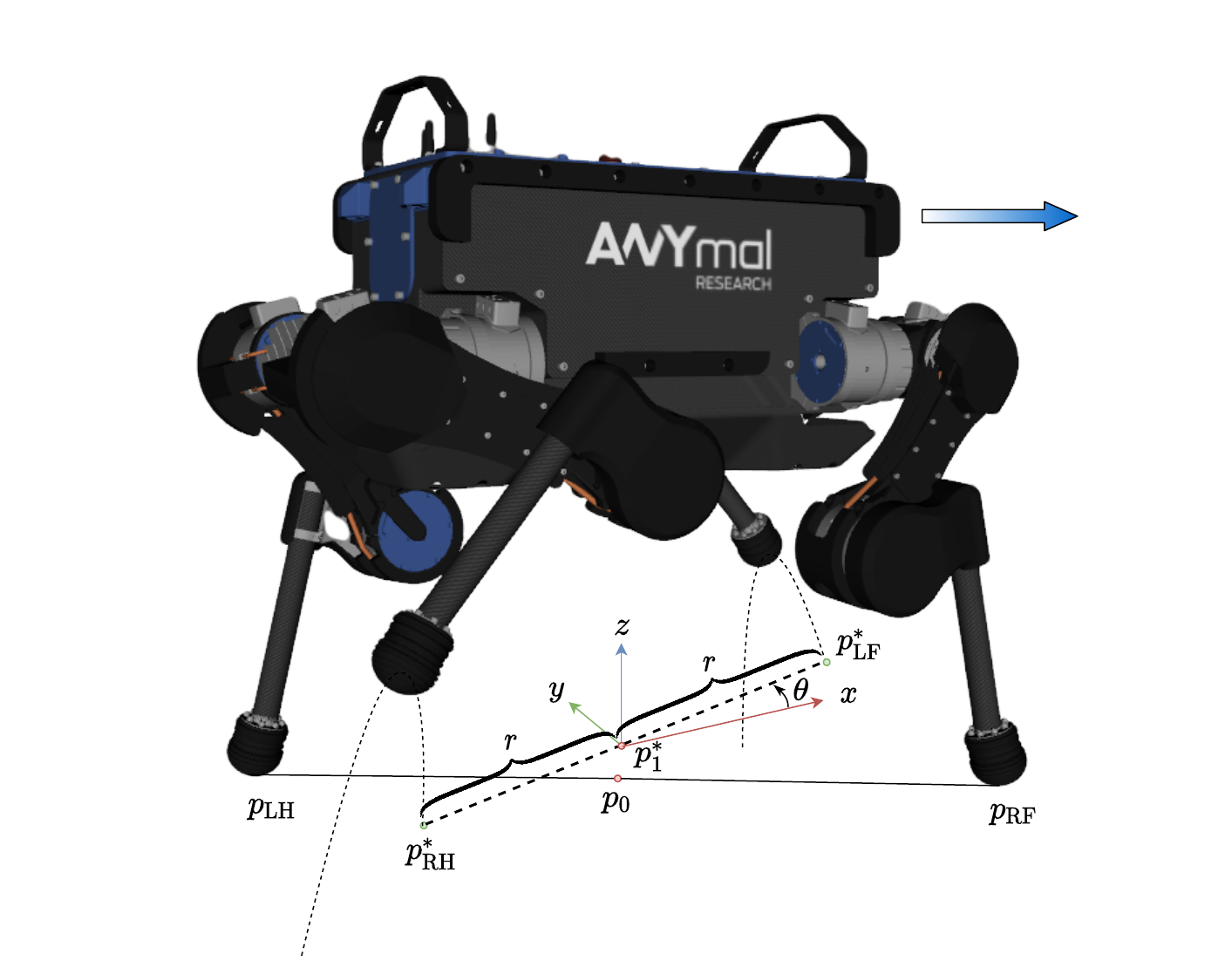}
 \caption{Geometrical relationship between footholds and  ZMPs. We assign the current ZMP ($p_0$) to be the middle point of the support line at the touchdown moment. 
 The desired footholds ($p_{\text{LF}}^*$, $p_{\text{RH}}^*$) are calculated from the desired ZMP ($p_1^*$) and two foot pair parameters $r$ and $\theta$. $r$ determines the distance between the foot pair and $\theta$ determines the orientation of the foot pair with respect to the robot heading direction.
 Nominal values are used for these two parameters. If there is a given steering command $\omega_z$, the orientation can be updated $\theta=\theta_{0}+\omega_z \cdot dt$. }
 \label{f:trajectory}
\end{figure}

This section explains the algorithms to generate the desired trajectories of swing feet and the CoM for trotting gait based on the results of the MPC.
The MPC provides the optimal ZMP that should be on the line connecting the next pair of support legs. We choose the ZMP to be the middle point of the support line for trotting gait. We keep the distance from the ZMP to each support foot location to be a fixed value $r$. Then we use the following equations to compute the desired footholds when the feet are swinging (in Fig. \ref{f:trajectory}):

\begin{equation}\label{e:footholds}
\begin{split}
    \text{LF}: \quad &\begin{bmatrix}p_x^{\text{LF}}\\p_y^{\text{LF}}\end{bmatrix}= \begin{bmatrix}p_{x_1}^{*}\\p_{y_1}^{*}\end{bmatrix}+r\begin{bmatrix}\cos(\theta_0+\Delta\theta)\\\sin(\theta_0+\Delta\theta)\end{bmatrix}\\
   \text{RH}:\quad &\begin{bmatrix}p_x^{\text{RH}}\\p_y^{\text{RH}}\end{bmatrix}= \begin{bmatrix}p_{x_1}^{*}\\p_{y_1}^{*}\end{bmatrix}+r\begin{bmatrix}-\cos(\theta_0+\Delta\theta)\\-\sin(\theta_0+\Delta\theta)\end{bmatrix} \\
   \text{RF}:\quad &\begin{bmatrix}p_x^{\text{RF}}\\p_y^{\text{RF}}\end{bmatrix}= \begin{bmatrix}p_{x_1}^{*}\\p_{y_1}^{*}\end{bmatrix}+r\begin{bmatrix}\cos(\theta_0-\Delta\theta)\\-\sin(\theta_0-\Delta\theta)\end{bmatrix} \\
   \text{LH}:\quad &\begin{bmatrix}p_x^{\text{LH}}\\p_y^{\text{LH}}\end{bmatrix}= \begin{bmatrix}p_{x_1}^{*}\\p_{y_1}^{*}\end{bmatrix}+r\begin{bmatrix}-\cos(\theta_0-\Delta\theta)\\\sin(\theta_0-\Delta\theta)\end{bmatrix}
\end{split}
\end{equation}
where LF, RH, RF and LH are the abbreviations for left-fore, right-hind, right-fore and left-hind feet. $\theta_0$ is a constant angle measured in the default configuration while $\Delta\theta$ is the rotation command sent by the users. For pacing gait, $\theta_0=0$. 

Here we do not tackle the height changing issue. We use the current height of support feet to be the desired height of desired footholds for swing feet. The peak height during swing is a fixed relative offset. This technique has to be adapted for some tasks such as climbing stairs. However the robustness of the planner and controller can handle slightly rough terrains, as we demonstrate through experiments. After determining the desired footholds, we use cubic splines to interpolate the trajectories between the initial foot positions and desired footholds for the swing feet, and feed the one forward time step positions, velocities and accelerations to the controller.

The desired positions and velocities for CoM are determined by the LIPM, i.e., Eq. (\ref{e:LIPM}) where the initial states $\mathbf{x}_{CoM}^0$ and $\mathbf{y}_{CoM}^0$ are updating with 400 Hz as well. Setting the variable $t$ in Eq. (\ref{e:LIPM}) to be a constant value $t=2.5$ ms results in the desired CoM positions and velocities along $x$ and $y$ for controller. We set the desired height of CoM to be a constant value with respect to the average height of the support feet. 

\section{LQR for base control}
\label{section:lqr_for_base_control}

We continue to use our projected inverse dynamic control framework \cite{xin2020optimization} as it allows us to focus on designing trajectory tracking controllers without considering physical constraints. The physical constraints will be satisfied in an orthogonal subspace. This framework gives us the opportunity to use the classical LQR without any adaptation. 

The dynamics of a legged robot can be projected into two orthogonal subspaces by using the projection matrix ${\mathbf{P}=\mathbf{I}-\mathbf{J}_c^{+}\mathbf{J}_c}$ \cite{Aghili2005}\cite{mistry2012operational} as follows:

Constraint-free space:
\begin{equation}\label{e:PSpace}
\mathbf{P}\mathbf{M}\ddot{\mathbf{q}}+\mathbf{Ph}=\mathbf{PS}\boldsymbol{\tau}
\end{equation}

Constraint space:
\begin{equation}\label{e:orthoSpace}
(\mathbf{I}-\mathbf{P})(\mathbf{M}\ddot{\mathbf{q}}+\mathbf{h})=(\mathbf{I}-\mathbf{P})\mathbf{S}\boldsymbol{\tau}+\mathbf{J}_c^{\top}\boldsymbol{\lambda}_c
\end{equation}

where $\mathbf{q}=\begin{bmatrix}{}_{I}\mathbf{x}_b^\top &\mathbf{q}_j^\top\end{bmatrix}^\top \in SE(3) \times \mathbb{R}^n$, where ${}_{I}\mathbf{x}_b \in SE(3)$ denotes the floating base's position and orientation with respect to a fixed inertia frame $I$, meanwhile $\mathbf{q}_j \in \mathbb{R}^n$ denotes the vector of actuated joint positions. Also, we define the generalized velocity vector as ${\dot{\mathbf{q}}=\begin{bmatrix}{}_{I}\mathbf{v}_b^\top & {}_{B}\boldsymbol{\omega}_b^\top &\dot{\mathbf{q}}_j^\top\end{bmatrix}^\top \in \mathbb{R}^{6+n}}$, where ${}_{I}\mathbf{v}_b \in \mathbb{R}^3$ and ${}_{B}\boldsymbol{\omega}_b \in \mathbb{R}^3$ are the linear and angular velocities of the base with respect to the inertia frame expressed respectively in the $I$ and $B$ frame which is attached on the floating base. $\mathbf{M} \in \mathbb{R}^{(n+6)\times(n+6)}$ is the inertia matrix, $\mathbf{h} \in \mathbb{R}^{n+6}$ is the generalized vector containing Coriolis, centrifugal and gravitational effects,
$\boldsymbol{\tau} \in \mathbb{R}^{n+6}$ is the vector of torques, $\mathbf{J}_c \in \mathbb{R}^{3k\times(n+6)}$ is the constraint Jacobian that describes $3k$ constraints, $k$ denotes the number of contact points accounting foot contact and body contact,
$\boldsymbol{\lambda}_c \in \mathbb{R}^{3k}$ are constraint forces acting on contact points, and
\begin{equation}\label{S}
    \mathbf{S}=\begin{bmatrix}
        \mathbf{0}_{6 \times 6} & \mathbf{0}_{6 \times n}   \\
        \mathbf{0}_{n \times 6}   & \mathbf{I}_{n \times n}
    \end{bmatrix}
\end{equation}
is the selection matrix with $n$ dimensional identity matrix $\mathbf{I}_{n \times n}$.

Note that Eq. (\ref{e:PSpace}) together with Eq. (\ref{e:orthoSpace}) provides the whole system dynamics. The sum of the torque commands generated in the two subspaces will be the final command sent to the motors as shown in Fig. \ref{f:framework}. In this paper, we focus on trajectory tracking control in the constraint-free subspace. We refer to our previous paper \cite{xin2020optimization} for the inequality constraint satisfaction in the constraint subspace. The swing legs are controlled by impedance controllers proposed in our former paper \cite{xin2020optimization}. In this paper, we propose to replace the impedance controller for base control with an LQR controller, benefiting from the optimal gain matrix instead of the hand-tuned diagonal impedance gain matrices. 

The similar works of \cite{mason2014full}\cite{mason2016balancing} did not enforce any inequality constraints with the classical LQR controller. The advantage of using projected inverse dynamics is that we can satisfy hard constraints, such as torque limits and friction cone constraints, in the constraint space by solving a QP as shown in Fig. \ref{f:framework}, in case the LQR controller and impedance controller generate torque commands that violate those inequality constraints.

\subsection{Linearization in Cartesian space}
\label{subsection:linearization_in_cartesian_space}

Based on Eq. (\ref{e:PSpace}), we derive the forward dynamics 
\begin{equation}\label{e:acceleration}
    \ddot{\mathbf{q}}=\mathbf{M}_c^{-1}(-\mathbf{Ph}+\dot{\mathbf{P}}\dot{\mathbf{q}})+\mathbf{M}_c^{-1}\mathbf{PS}\boldsymbol{\tau}
\end{equation}{}
where $\mathbf{M}_c=\mathbf{PM}+\mathbf{I}-\mathbf{P}$ is called constraint inertia matrix \cite{Aghili2005}. Eq. (\ref{e:acceleration}) could be linearized with respect to the full state vector $\begin{bmatrix}\mathbf{q}^\top &\dot{\mathbf{q}}^\top\end{bmatrix}^\top$. However, the resulting linearized system would not be controllable as the corresponding controllability matrix is not full rank. Instead of resorting to one more projection as done in \cite{mason2014full}, we linearize the dynamics in the Cartesian space to control only the base states rather than all the states of a whole robot.

Just using a selection matrix, we can derive the forward dynamics with respect to ${}_{I}\mathbf{x}_b$
\begin{equation}\label{e:baseDynamics}
    \ddot{\mathbf{x}}_b=\mathbf{J}_b\mathbf{M}_c^{-1}(-\mathbf{Ph}+\dot{\mathbf{P}}\dot{\mathbf{q}})
    +\mathbf{J}_b\mathbf{M}_c^{-1}\mathbf{PS}\boldsymbol{\tau}=f({}_{I}\mathbf{x}_b,\mathbf{\dot{x}}_b,\boldsymbol{\tau})
\end{equation}
where $\mathbf{J}_b=\begin{matrix}[\mathbf{I}_{6\times6} &\mathbf{0}_{6\times n}]\end{matrix}_{6\times(n+6)}$, $\dot{\mathbf{x}}_b=\begin{bmatrix}{}_{I}\mathbf{v}_b^\top & {}_{B}\boldsymbol{\omega}_b^\top \end{bmatrix}^\top$. 

By using Euler angles for the orientation in ${}_{I}\mathbf{x}_b$, we can define the state vector as 
\begin{equation}
    \mathbf{X}=\begin{bmatrix}
    {}_{I}\mathbf{x}_b\\
    \dot{\mathbf{x}}_b
    \end{bmatrix}_{12 \times 1}
\end{equation}{}
We linearize Eq. (\ref{e:baseDynamics}) to state space dynamics around a configuration $(\mathbf{q}_0, \dot{\mathbf{q}}_0, \boldsymbol{\tau}_0)$ where $\boldsymbol{\tau}_0$ is the gravity compensation torques, yielding
\begin{equation}\label{e:baseStateEquation}
    \dot{\mathbf{X}}=\mathbf{A}_0^b\mathbf{X}+\mathbf{B}_0^b\boldsymbol{\tau}
\end{equation}
Eq. (\ref{e:baseStateEquation}) is detailed as 
\begin{equation}
    \begin{bmatrix}
    \dot{\mathbf{x}}_b\\
    \ddot{\mathbf{x}}_b
    \end{bmatrix}=\begin{bmatrix}
    \mathbf{0} & \mathbf{I}\\
    \mathbf{A}_{21} & \mathbf{A}_{22}
    \end{bmatrix}\begin{bmatrix}
    {}_{I}\mathbf{x}_b\\
    \dot{\mathbf{x}}_b
    \end{bmatrix}+\begin{bmatrix}
    \mathbf{0}\\
    \mathbf{B}_2
    \end{bmatrix}\boldsymbol{\tau}
\end{equation}{}
where $\mathbf{A}_{21}$, $\mathbf{A}_{22}$ and $\mathbf{B}_{2}$ are defined as
\begin{equation}\label{e:A21}
    \mathbf{A}_{21}=\frac{\partial f({}_{I}\mathbf{x}_b,\mathbf{\dot{x}}_b,\boldsymbol{\tau})}{\partial {}_{I}\mathbf{x}_b}|_{\mathbf{q}_0, \dot{\mathbf{q}}_0, \boldsymbol{\tau}_0}
\end{equation}

\begin{equation}\label{e:A22}
    \mathbf{A}_{22}=\frac{\partial f({}_{I}\mathbf{x}_b,\mathbf{\dot{x}}_b,\boldsymbol{\tau})}{\partial \dot{\mathbf{x}}_b}|_{\mathbf{q}_0, \dot{\mathbf{q}}_0}
\end{equation}

\begin{equation}
    \mathbf{B}_2=\frac{\partial f({}_{I}\mathbf{x}_b,\mathbf{\dot{x}}_b,\boldsymbol{\tau})}{\partial \boldsymbol{\tau}}=\mathbf{J}_b\mathbf{M}_c^{-1}\mathbf{PS}|_{\mathbf{q}_0}
\end{equation}

For simplicity, we use a central finite difference method to compute the partial derivatives of Eq. (\ref{e:A21}) and Eq. (\ref{e:A22}). The deviation factor for finite difference we used for the experiments is $\num{1e-5}$. 

\subsection{LQR controller}
\label{subsection:lqr_controller}

We consider Eq. (\ref{e:baseStateEquation}) as a linear time-invariant system and solve the infinite horizon LQR problem to compute the optimal feedback gain matrix $\mathbf{K}$. The cost function to be minimized is defined as
\begin{equation}\label{e:LQR}
    J=\int_0^\infty({\mathbf{X}}^\top\mathbf{Q}\mathbf{X}+\boldsymbol{\tau}^\top \mathbf{R}\boldsymbol{\tau})dt
\end{equation}
and the resulting controller for the base control is
\begin{equation}\label{e:LQRcommands}
    \boldsymbol{\tau}_{m_2}=\mathbf{K}(\mathbf{X}_d-\mathbf{X})+\boldsymbol{\tau}_d
\end{equation}
where $\mathbf{X}_d$ is the desired state, $\boldsymbol{\tau}_d$ is the feedforward term derived from inverse dynamics based on the desired state.

We use ADRL Control Toolbox (CT) \cite{adrlCT} to solve the infinite-horizon LQR problem and obtain the $\mathbf{K}$ matrix. It should be noted that the linearization is computed in every control cycle based on the current configuration $(\mathbf{q}_0, \dot{\mathbf{q}}_0, \boldsymbol{\tau}_0)$. The $\mathbf{K}$ matrix is updated at 400 Hz, which is different to \cite{mason2016balancing} where they only compute the $\mathbf{K}$ matrices corresponding to few key configurations. We think linearization should be updated around current configuration in order to improve computation accuracy if the computation is fast enough.

In practice, we increase the weights in $\mathbf{R}$ of Eq. (\ref{e:LQR}) for swing legs, relying more on the support legs for base control. Otherwise, the torque commands of Eq. (\ref{e:LQRcommands}) can affect the tracking of swing trajectories too much. 

In addition, the motion planner in Section \ref{section:motion_generation} feeds the desired CoM trajectory to the controllers, whereas the LQR controller controls the base pose. In theory, we should replace ${}_{I}\mathbf{x}_b$ with $\mathbf{x}_{CoM}$ in Eqs. (\ref{e:PSpace})(\ref{e:orthoSpace}) and transform the dynamic equations to be with respect to CoM variables as in \cite{henze2016passivity}. Then the LQR controller will directly track the desired CoM trajectory. In this paper, we approximately consider the translation of base along $x$ and $y$ aligned with CoM since the base dominates the mass of the whole robot. 

\section{Validations}
\label{section:experiments}

We use a torque controllable quadruped robot ANYmal made by ANYbotics to conduct our experiments. The onboard computer has an Intel 4th generation (HaswellULT) i7-4600U (1.4 GHz-2.1 GHz) processor and two HX316LS9IBK2/16 DDR3L memory cards. The robot weights approximately $\SI{35}{\kilogram}$ and has 12 joints actuated by Series Elastic Actuators (SEAs) with maximum torque of \SI[inter-unit-product =\ensuremath{\cdot}]{40}{\newton\meter}. The real-time control cycle is $\SI{2.5}{\milli\second}$. The control software is developed based on Robot Operating System 1 (ROS 1). We use the dynamic modeling library Pinocchio \cite{carpentier2019pinocchio} to perform the model linearization of Section \ref{section:lqr_for_base_control}. An active set method based QP solver provided by ANYbotics is used to solve the QPs for the MPC planner and the controller. A video of the experimental results can be found at: \url{https://youtu.be/khP6PQ9xuso}.

\begin{figure}[b!]
\centering
 \includegraphics[width=\linewidth]{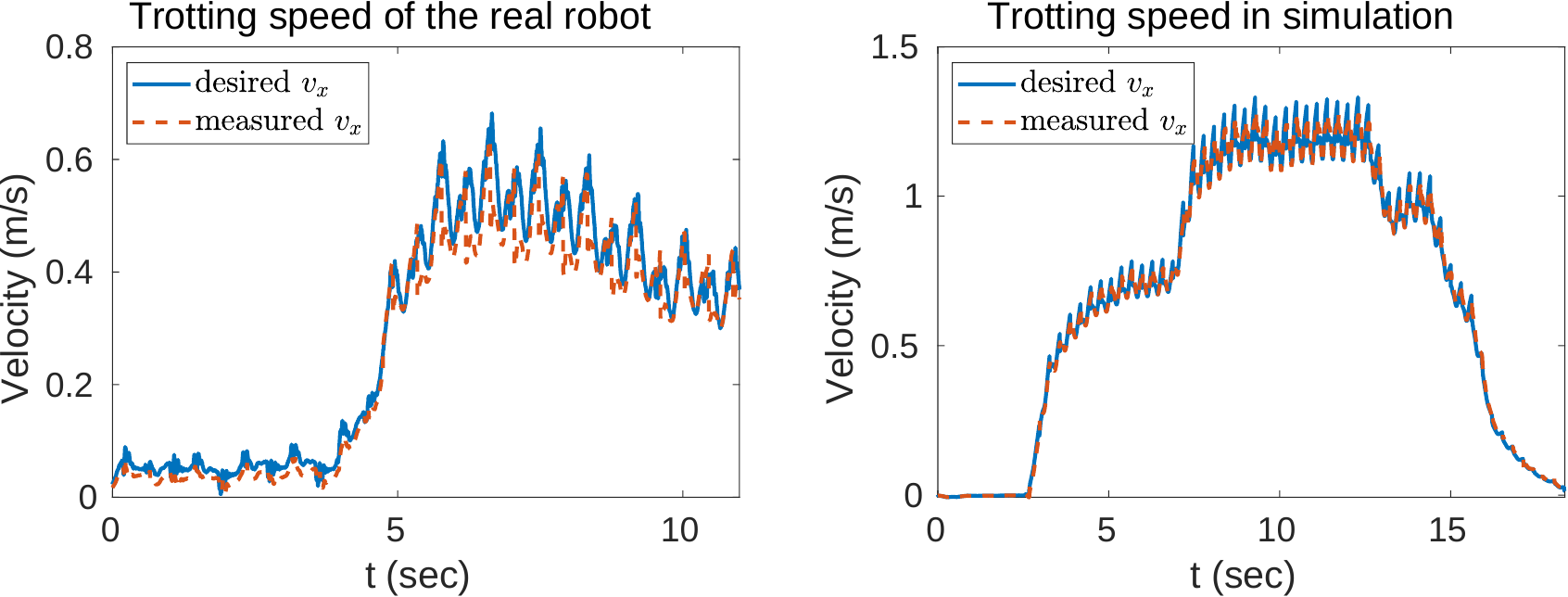}
 \caption{Recorded fastest trotting speeds on real robot and in simulation. The desired velocities are generated from LIPM dynamics, i.e. Eq. (\ref{e:LIPM}), which explains why the reference velocities are not smooth.}
 \label{f:fastestSpeed}
\end{figure}

\begin{figure*}[t!]
\centering
\includegraphics[width=\linewidth]{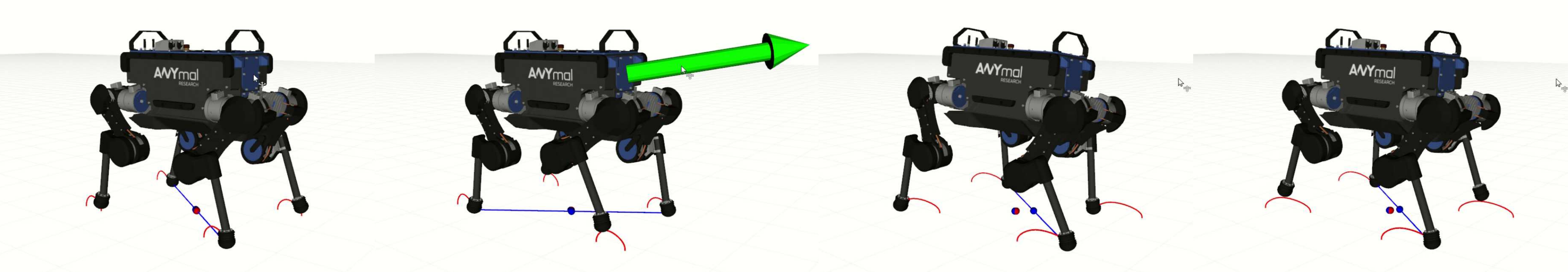}
\caption{Footstep and swing trajectory replanning under disturbances. The robot is walking forward and an external force (green arrow) is applied to the base of the robot. The push results in a sudden change of the CoM position and velocity. The footstep planner uses the updated state to replan the footholds. The swing trajectories (red lines) are updated accordingly.}
\label{f:pushSim} 
\end{figure*}

\subsection{Trotting speed}
\label{subsection:trotting_speed}

We first tested the fastest walking speed when using the proposed algorithms. Figure \ref{f:fastestSpeed} shows the recorded speeds along $x$ direction in real robot experiment and in simulation. In simulation, the robot could stably trot forward with maximum speed 1.2 m/s. On real robot, the maximum speed reached 0.6 m/s. The results are reasonable since the trotting gait does not have a flying phase. The fact that the real robot cannot achieve as fast motion as in simulation is also reasonable considering model errors and other uncertainties. Model errors also cause drifting on the real robot which is difficult to resolve without external control loops. Constant values for the parameters of the gait planner were employed. They are $T_s=0.3$, $z_{CoM}=0.42$, $g=-9.8$, $N=3$, $Q_i=1000$, $R_i=1$, $\theta_0=0.56$, $r=0.41$. It should be noted that the prediction number $N$ in the MPC does not need to be as large as possible with the concern of computation efficiency. We tested $N=2 \sim 5$, and they showed similar performance.

\subsection{Push recovery}
\label{subsection:pushing_recovery}

\begin{figure}[t!]
\includegraphics[width=\linewidth]{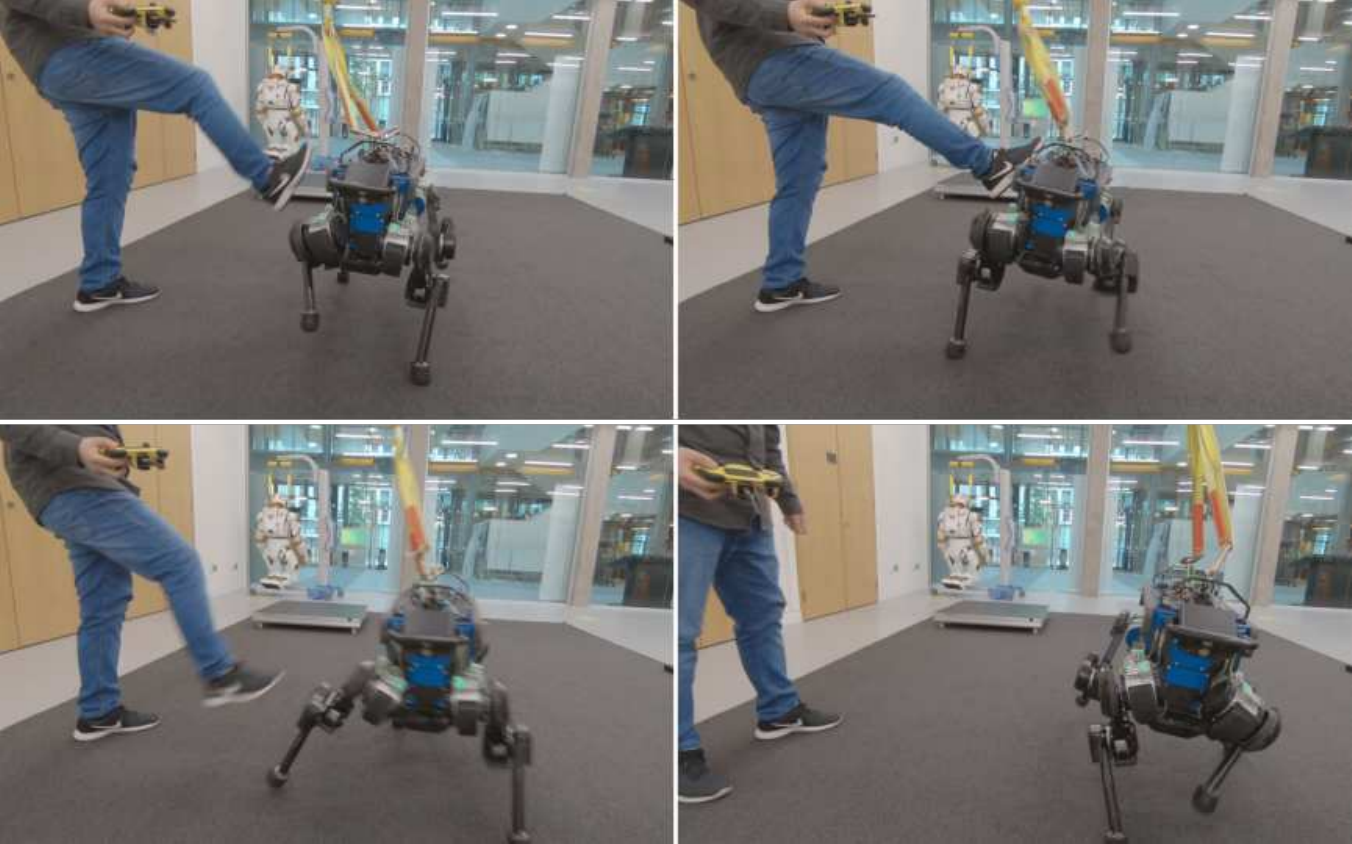}
 \caption{Kicking the robot during trotting. Benefiting from the 400 Hz MPC update frequency, the robot can quickly update the optimal footholds to recover from disturbances.}
 \label{f:pushReal}
\end{figure}

\begin{figure}[t!]
\centering
 \includegraphics[width=\linewidth]{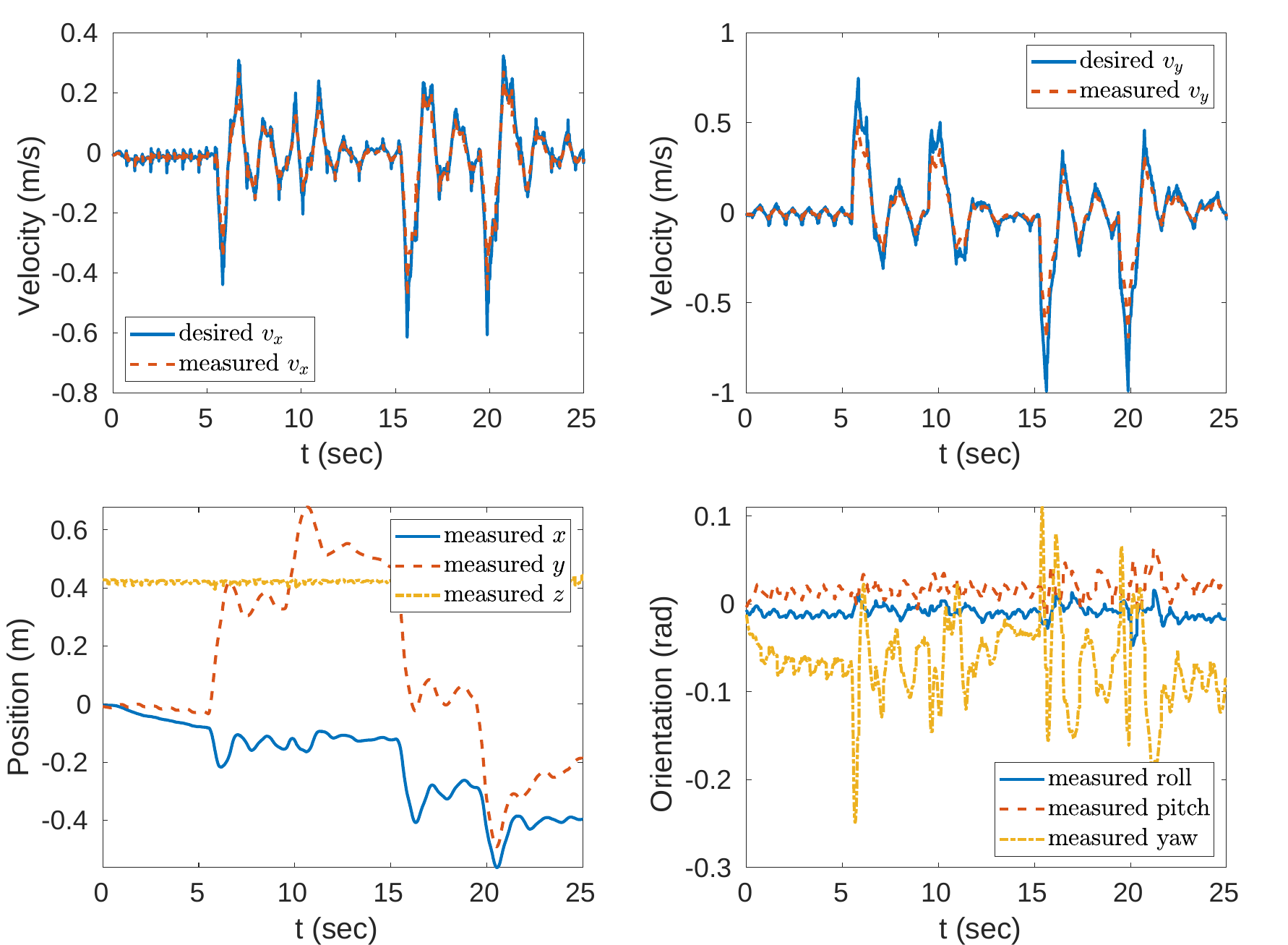}
 \caption{Recorded state when the robot was kicked four times. The desired Euler angles were 0. The robot was quickly regulated back to normal even though the velocity reached $\SI{-1}{\meter/\second}$ after kicking.}
 \label{f:pushRecoveryData}
\end{figure}

In this subsection, we demonstrate the benefit of high frequency replanning for disturbance rejection. We first use simulation to show the replanned footholds and trajectories as shown in Fig. \ref{f:pushSim}. The disturbance is added when RF and LH feet are swinging. The disturbance results in sharp state changes. The MPC computed the new footholds after receiving the updated state. Figure \ref{f:pushReal} presents snapshot photos of the push-recovery experiment on the real robot during trotting while recorded state data is shown in Fig. \ref{f:pushRecoveryData}. The robot was kicked four times roughly along the $y$ direction. We can see the peak velocity of $y$ reached $\SI{-1}{\meter/\second}$ during the last two kicks, but it was quickly regulated back to normal using one or two steps. The orientation did not change too much after kicking, which also indicates the robustness of the method.

\begin{figure}[t!]
\centering
 \includegraphics[width=\linewidth]{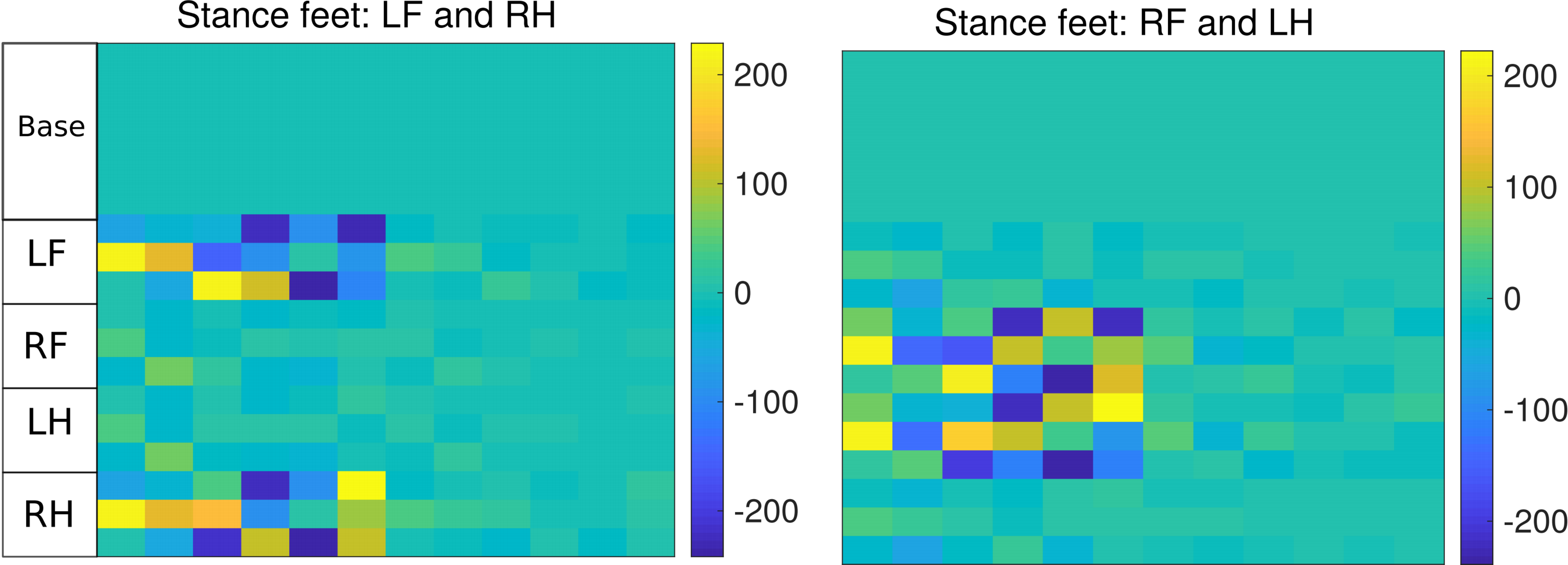}
 \caption{Visualization of the gain matrix as computed by the LQR controller during trotting. The size of the gain matrix is 18 by 12. The first 6 columns correspond to the position and orientation while the remaining 6 columns are for velocity control. Off-diagonal gains demonstrate that dynamic coupling effects may be exploited for control.}
 \label{f:K}
\end{figure}

\subsection{Balance control}
\label{subsection:balance_control}

Most of the trotting gait control algorithms rely on quick switching of swing and stance phases to achieve dynamic balance. Recently researchers demonstrated that quadruped robots with point feet can stand on two feet to maintain balance \cite{chignoli2020variational}\cite{gonzalez2020line}. Although we did not manage balancing on two feet on our robot, we compared the longest period of swing phase of trotting when using our proposed LQR controller versus the default trotting controller of ANYmal \cite{gehring2013control}. The longest swing phase when using LQR is $\SI{0.63}{\second}$ whereas the default controller only achieved $\SI{0.42}{\second}$. When the base is controlled by our previous impedance controller, the longest swing phase is $\SI{0.4}{\second}$. This verifies the improved performance of our LQR controller in terms of balance control. Figure \ref{f:K} shows two gain matrices for the two phases of trotting during this experiment. It should be noted that the gain matrices are updated in every control cycle (but the changes are small). We can see that the elements of the first 6 rows are 0 because of the existence of selection matrix $\mathbf{S}$. The $\mathbf{Q}_{12 \times 12}$ we used in the experiment was $\mathbf{Q}=\text{diag}(\text{diag}(1500)_{6 \times 6}, \enskip \text{diag}(1)_{6 \times 6})$. The $\mathbf{R}_{18 \times 18}$ was switched depending on the phase. The diagonal elements corresponding to the two swing legs in $\mathbf{R}$ are 10 times larger than the other diagonal elements for stance legs and the base, reducing the efforts of swing legs in balance control. The elements in $\mathbf{R}$ for stance legs and the base we used in this experiment are 0.03.

\subsection{Trotting on slippery terrains}
\label{subsection:trotting_on_slippery_terrains}

\begin{figure}[t!]
\centering
 \includegraphics[width=\linewidth]{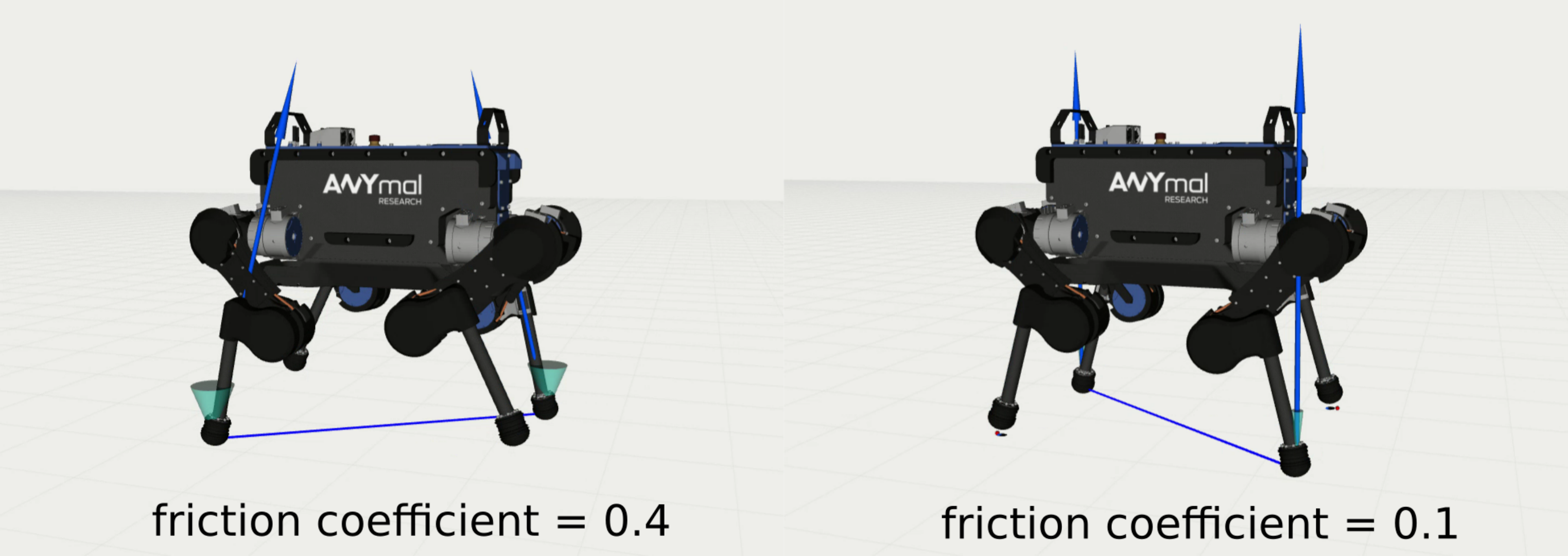}
 \caption{The friction cone constraints are satisfied by the controller. The blue arrows represent the actual contact forces while the green cones denote the friction cones.}
 \label{f:friction}
\end{figure}

The most important advantage of the proposed control framework compared to similar works \cite{mason2014full}\cite{mason2016balancing} is that we can satisfy inequality constraints while using LQR for trajectory tracking. We do not change the classical LQR to be a constrained LQR. By contrast, using the projected inverse dynamic control allows us to satisfy inequality constraints in the constraint subspace. The LQR controller only serves as a trajectory tracking controller and does not need to consider the inequality constraints. The QP optimization in the constraint subspace plays the role of trading off different constraints. For example, as we have shown in our previous paper \cite{xin2020optimization} for static gait, trajectory tracking performance will be sacrificed to prevent slipping if torque commands for trajectory tracking generate contact forces beyond the friction cones. Here we demonstrate that our proposed controller can satisfy friction cone constraints for dynamic gaits as well. Figure \ref{f:friction} shows the controller can keep the contact forces within the friction cone after reducing the friction coefficient to match the actual friction coefficient of the terrain. The smallest friction coefficient we achieved in simulation for trotting in spot on flat terrain is 0.07. However it is difficult to trot on such slippery terrain because the trajectory tracking is quite poor in this situation.

\subsection{Transition from trotting to pacing}
\label{subsection:transition_from_trotting_to_pacing}

\begin{figure}[t!]
\centering
 \includegraphics[width=\linewidth]{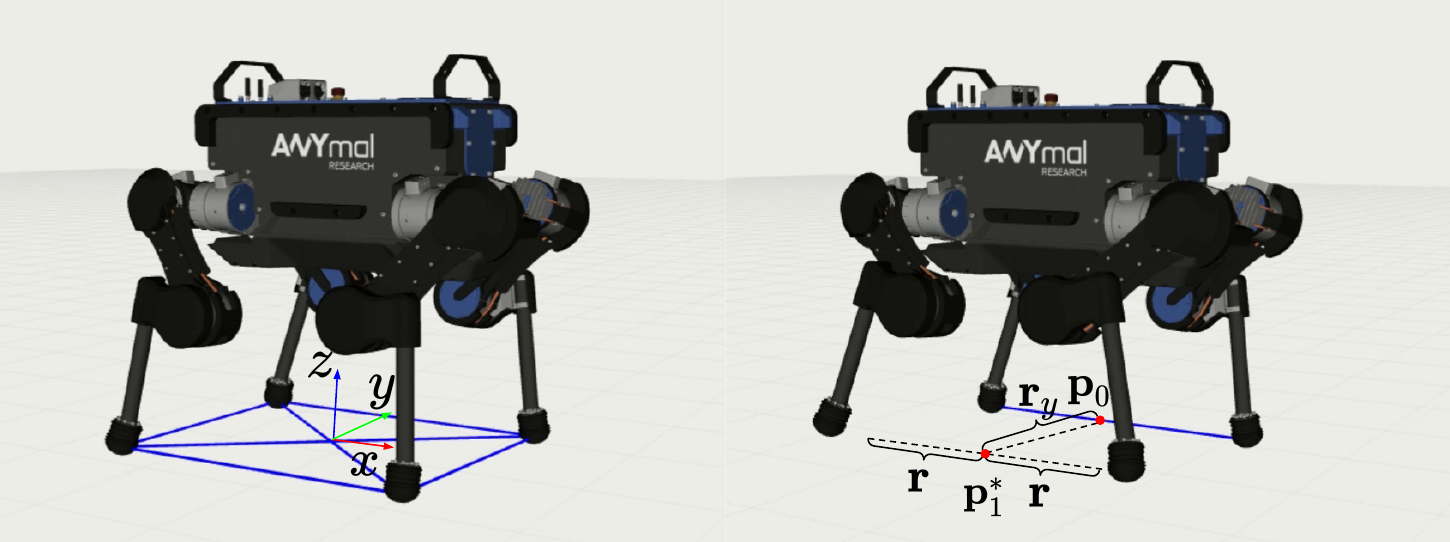}
 \caption{A base shifting motion is needed to transit from trotting to pacing.}
 \label{f:trotting2pacing}
\end{figure}

Pacing gait is a more dynamic gait compared to trotting since the CoM is always off the supporting line. The difference between trotting and pacing in terms of the MPC formulation is that there will be a constant offset $r_y$ between $p_{y_i}$ and $p_{y_{i-1}}$ (see Fig. \ref{f:trotting2pacing}) in the cost function Eq. (\ref{e:MPCCostY}) in order to avoid conflicts of the right and left feet. In our experiment, we specified a transition motion of shifting the base to a side to start pacing. We can also remove this transition motion by reducing the gait period or reducing the distance between left and right feet. The gait period in this experiment was $\SI{0.44}{\second}$ with $r_y=\SI{0.08}{\meter}$. On the controller side, we used the same $\mathbf{Q}$ and $\mathbf{R}$ for trotting and pacing.

\subsection{Outdoor test with adaptive feet}
\label{subsection:traversing_terrains_with_gravels}
\begin{figure}[t]
\centering
\includegraphics[width=\linewidth]{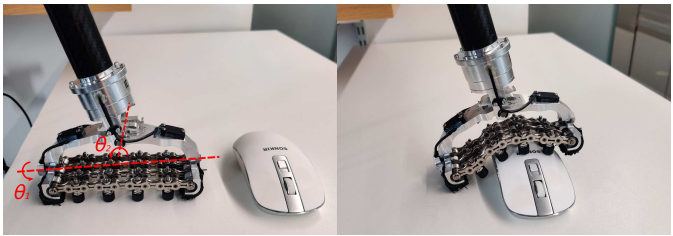}
\caption{The SoftFoot-Q, an adaptive foot for quadrupeds. $\theta_1$ and $\theta_2$ indicate the passive joints of the ankle.}
\label{f:flexibleFeet} 
\end{figure}

\begin{figure}[t]
\centering
 \includegraphics[width=\linewidth]{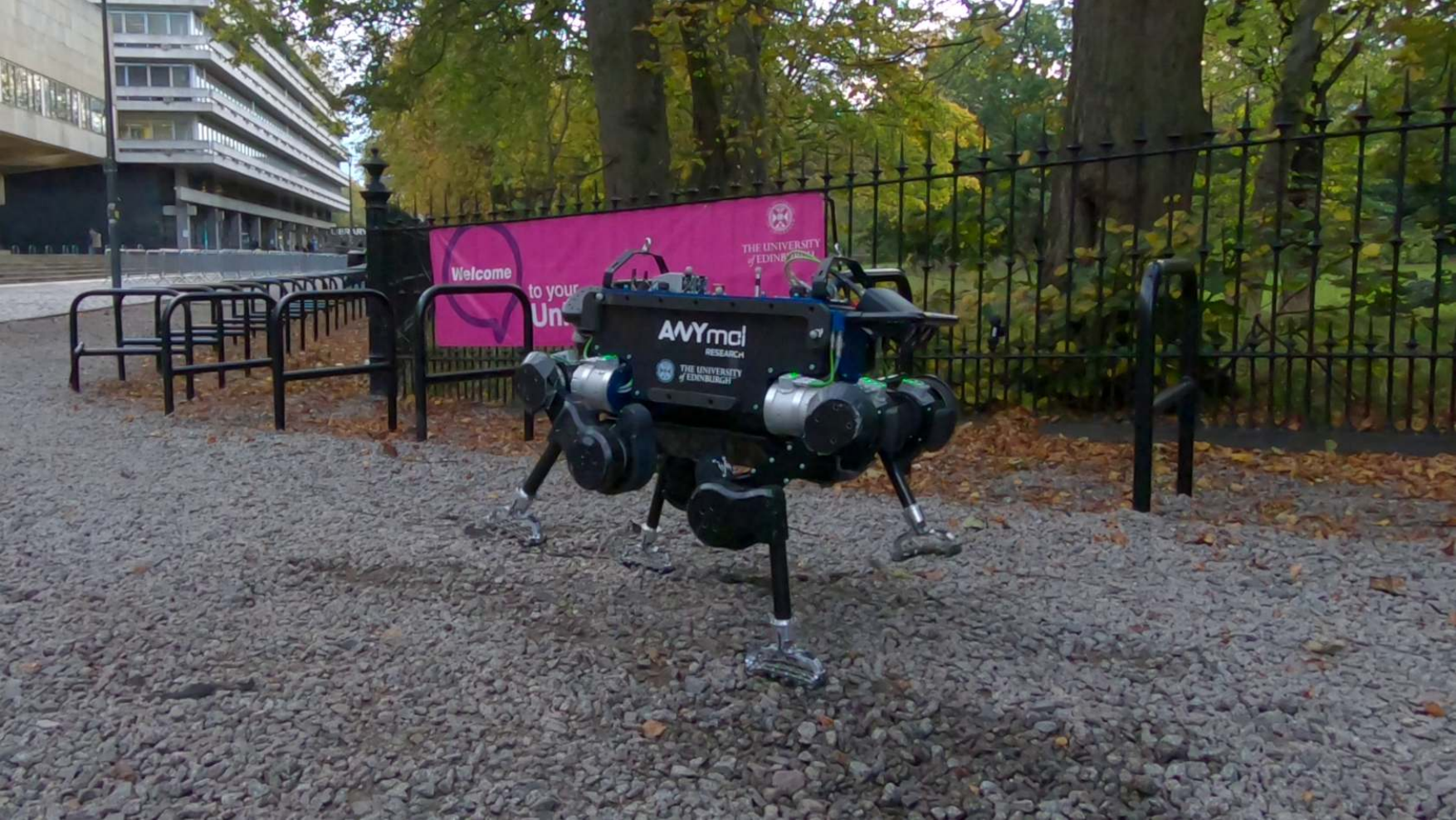}
 \caption{Trotting out of the lab with adaptive feet on rubble terrain. }
 \label{f:outside1}
\end{figure}

In this subsection, we test the versatility of our approach with adaptive feet SoftFoot-Q \cite{adaptivefeet2021} in outdoor environments. Figure \ref{f:flexibleFeet} shows a typical case of the adaptive feature. Compared to the traditional sphere feet, the adaptive feet have larger contact surface. Those features will benefit the traversability of rough
terrains with rocks, loose gravel and rubble by enlarging the
contact surfaces with ground. We performed experiments in trotting locomotion on rough terrains outside our lab as shown in Fig. \ref{f:outside1}.
It should be noted that our controller did not take the two DOF (one DOF less than the case with a spherical foot) passive ankle into account. The model errors caused by the adaptive feet were treated as disturbances by the controller, where the success of the tests shows the robustness of our controller.

\section{CONCLUSIONS}
\label{section:conclusions}
This paper presents a full control framework for dynamic gaits where all the modules are running with the same frequency. The robustness of the dynamic walking is improved significantly by two factors. The first factor is the MPC planner, which mostly contributes to rejecting large disturbances, such as kicking the robot, because the MPC uses footsteps to regulate the state of the robot. The second factor is the LQR controller for balancing control, which also undertakes the duty of trajectory tracking. The method is general and shown to able to work both with spherical and adaptive feet. The latter were seen to reduce the slipping chance on rough terrains. The outdoor experiments demonstrate the robustness of locomotion after adopting the proposed methods and assembling the adaptive feet. 

Future work will focus on adapting the current planner to consider terrain information to handle large slopes and stairs. Also, the new feet can be used to measure the local inclination of the ground which can improve the accuracy of the terrain information, similar to \cite{8968320}.






\section*{ACKNOWLEDGMENT}

The authors would like to thank Dr. Quentin Rouxel and Dr. Carlos Mastalli for introduction on using the Pinocchio rigid body dynamics library. The authors also would like to thank the editor and reviewers for their useful comments.


\bibliographystyle{IEEEtran}
\bibliography{IEEEabrv,bibliography_abrv}

\end{document}